\documentclass[10pt,twocolumn,letterpaper]{article}

\usepackage[pagenumbers]{cvpr} 

\usepackage{ragged2e}
\usepackage{graphicx}
\usepackage{amsmath}
\usepackage{amssymb}
\usepackage{booktabs}
\usepackage{multirow} 
\usepackage{makecell}
\usepackage{colortbl}
\usepackage[table]{xcolor}
\usepackage[pagebackref,breaklinks,colorlinks]{hyperref}

\usepackage[capitalize]{cleveref}
\crefname{section}{Sec.}{Secs.}
\Crefname{section}{Section}{Sections}
\Crefname{table}{Table}{Tables}
\crefname{table}{Tab.}{Tabs.}

\def\cvprPaperID{2130} 
\def\confName{CVPR}
\def\confYear{2024}

\begin{document}

\title{\textsc{CDAD-Net:} Bridging Domain Gaps in Generalized Category Discovery}

\author{Sai Bhargav Rongali$^{*1}$ \and Sarthak Mehrotra$^{*1}$ \and Ankit Jha $^{2}$\and Mohamad Hassan N C$^{1}$ \and Shirsha Bose$^{3}$ \and Tanisha Gupta$^{1}$ \and Mainak Singha$^{1}$ \and Biplab Banerjee$^{1}$
\and
$^{1}$Indian Institute of Technology Bombay, India \and
\and $^{2}$INRIA, Grenoble, France
$^{3}$Technical University of Munich, Germany
\and
{\tt\small \{rongalisaibhargav002, sarthak2002.mehrotra, ankitjha16, mohdhassannc, shirshabosecs,} \\ \tt\small{tanishagupta072, mainaksingha.iitb, getbiplab\}@gmail.com}}

\twocolumn[{%
\renewcommand\twocolumn[1][]{#1}%
\maketitle
\begin{center}
    \centering
    \captionsetup{type=figure}
    \vspace{-7mm}\includegraphics[width=0.75\textwidth,height=3cm]{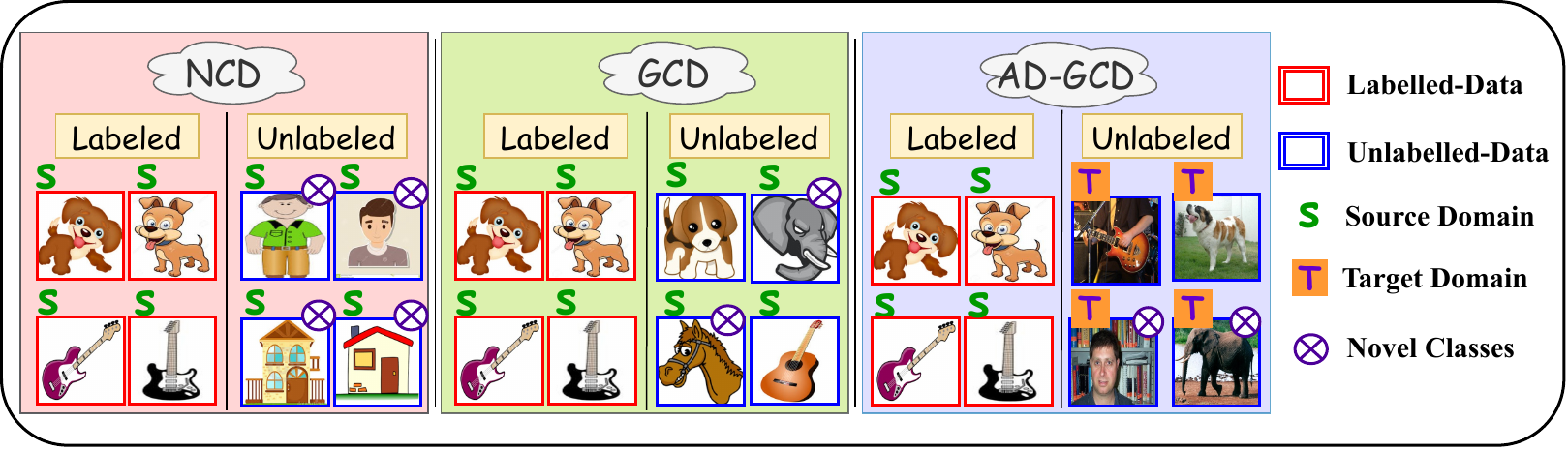}
    \vspace{-3.7mm}
    \captionof{figure}{We introduce the problem setting of \textbf{across domain generalized category discovery} (AD-GCD), which is different from the traditional GCD setting \cite{vaze2022generalized} in that we consider the labeled and unlabeled data to arise from different data distributions. We call the labeled data the \texttt{Source} domain and the unlabeled data the \texttt{Target} domain, respectively.}
    \label{fig:Teaser11}
    \vspace{-0.1cm}
\end{center}%
}]

\maketitle
\def\thefootnote{*}\footnotetext{Equal Contribution}\def\thefootnote{\arabic{footnote}}

\begin{abstract}
\vspace{-2mm}
In Generalized Category Discovery (GCD), we cluster unlabeled samples of known and novel classes, leveraging a training dataset of known classes. A salient challenge arises due to domain shifts between these datasets. To address this, we present a novel setting: Across Domain Generalized Category Discovery (AD-GCD) and bring forth \textsc{CDAD-Net} (\underline{C}lass \underline{D}iscoverer \underline{A}cross \underline{D}omains) as a remedy.
\textsc{CDAD-Net} is architected to synchronize potential known class samples across both the labeled (source) and unlabeled (target) datasets, while emphasizing the distinct categorization of the target data. To facilitate this, we propose an entropy-driven adversarial learning strategy that accounts for the distance distributions of target samples relative to source-domain class prototypes. Parallelly, the discriminative nature of the shared space is upheld through a fusion of three metric learning objectives.
In the source domain, our focus is on refining the proximity between samples and their affiliated class prototypes, while in the target domain, we integrate a neighborhood-centric contrastive learning mechanism, enriched with an adept neighbors-mining approach.
To further accentuate the nuanced feature interrelation among semantically aligned images, we champion the concept of conditional image inpainting, underscoring the premise that semantically analogous images prove more efficacious to the task than their disjointed counterparts.
Experimentally, \textsc{CDAD-Net} eclipses existing literature with a performance increment of 8-15\% on three AD-GCD benchmarks we present.
\end{abstract}

\section{Introduction}
Recent times have witnessed the rise of deep learning techniques for visual inference \cite{he2016deep, redmon2016you, vit}. However, their success predominantly relies on learning from ample supervision, which can be costly at times. 
One of the solutions is semi-supervised learning \cite{semisl}, which entails learning from both labeled and unlabeled sample sets within a closed-set framework, where both sets encompass identical categories. A more meaningful approach would be not to constrain the classes in the unlabeled set. Accordingly, the concept of novel category discovery (NCD) has recently been introduced \cite{zhong2021neighborhood, Fini_2021_ICCV} that aims to semantically group unlabeled data of unseen categories leveraging supervised knowledge from a distinct set of known categories.
However, NCD assumes that all unlabeled instances belong to novel categories with respect to the labeled data, which is often impractical. Vaze \etal \cite{vaze2022generalized} extended NCD to the GCD setting to address this limitation and allowed unlabeled images to originate from both known and novel classes.

A prevailing assumption in extant GCD methods \cite{vaze2022generalized, Zhang2022PromptCALCA} is that both the labeled and unlabeled datasets uniformly adhere to identical distributions. This assumption facilitates the smooth transfer of discriminative knowledge from labeled to unlabeled sets. Yet, in the real-world context, domain shifts are commonplace. Such shifts can result in the labeled and unlabeled data stemming from markedly different visual domains, impeding the seamless knowledge propagation between the datasets.

Building upon these discussions, this paper introduces a previously unexplored problem setting known as AD-GCD, summarized in Fig. \ref{fig:Teaser11}. In this context, we designate the domain possessing labeled data as the \textit{source} domain, while its unlabeled counterpart, possibly emanating from disparate data distributions and containing samples from both known and novel classes, is referred to as the \textit{target} domain. Employing a conventional GCD approach may fall short in addressing AD-GCD challenges, as the domain discrepancies could result in inaccurate mappings of novel-class samples from the target to known classes in the source domain, thereby detrimentally impacting performance.
Moreover, although augmenting a GCD solution with open-set domain adaptation (OSDA) \cite{saito2018open, rakshit2020multi} could enhance the mapping of known-class samples between domains, OSDA tends to hamper the cluster topology of the novel-class samples. The closest existing work to ours is \cite{yu2022self}, which focuses on clustering target-domain novel-class samples into a predetermined number of known clusters and classifying potential target samples that belong to training classes, using a classifier trained on the source domain. Conversely, our AD-GCD framework endeavors to cluster the target domain accurately without prior knowledge of cluster numbers, positioning our approach as orthogonal to the current relevant body of literature.

AD-GCD has many potential real-world applications.
 For example, autonomous vehicles navigating diverse environments, often confront unfamiliar objects or road conditions outside their training data. Initially, these models are trained using either synthetic data or a limited set of real-world data, while the actual deployment scenarios may vary significantly.
AD-GCD plays a pivotal role in this case by aiding in categorising these unfamiliar elements and recognising known events, enabling more effective handling of these situations.

Consequently, we assert two principal objectives for successfully addressing the challenges of AD-GCD:
\textbf{i)} Achieve alignment of the known set of classes across domains to felicitate discriminative knowledge transfer from the source to the target domain.
\textbf{ii)} Ensure that the domains remain discriminative with respect to both known and novel classes.

\noindent \textbf{The proposed solution:}  We present \textsc{CDAD-Net}, a novel framework to tackle AD-GCD. Towards solving our primary objective, we advocate for a cross-domain alignment, deploying a novel entropy optimization strategy. This is anchored on assessing the similarity distributions of target samples vis-à-vis the class prototypes of the source domain.
Underpinning our approach is the theory that target samples aligning with the source domain would show a decline in entropy of the aforesaid distribution, each trending towards a particular prototype. Conversely, samples representing novel classes are projected to display increased entropy of the distribution, highlighting their departure from established class boundaries.
Clearly, our approach takes a different path from traditional OSDA practices. Rather than funneling target novel-class samples into one cohesive cluster, our technique deliberately sidesteps this convention.

To achieve our secondary objective of amplifying the distinctiveness within the shared embedding space, we scrutinize existing techniques. Established GCD methodologies \cite{pu2023dynamic, vaze2022generalized} advocate for an augmentation-based contrastive self-supervision method like \cite{SimCLR2020a} across both labeled and unlabeled data for representation learning. Nonetheless, as reported in \cite{kalibhat2022multidomain}, such techniques work well for a single domain, but are unfavorable for multi-domain data.

This drives our pivot towards accentuating domain-specific discriminative features, where we propose to harness neighborhood-centric contrastive learning. Specifically for the unlabeled target domain, we offer to reuse the aforementioned distance dynamics interlinking target samples and class prototypes from the source domain to obtain the positive and negative neighbors per reference image. 

Our existing global image embedding-based losses might inadequately capture local image variations per category for different visual domains. To enhance these embeddings with detailed, local attributes for better clustering, we introduce the task of \textit{conditional image inpainting}. In this approach, we base the patch-level reconstruction of an image on another complete image. We hypothesize that a conditioning image with semantically similar traits to the reference will yield superior results compared to one with divergent semantics. This concept is operationalized using a newly developed loss function.

The culmination is a meticulously aligned and discriminative embedding space, paving the way for superior cluster assignment of target domain samples via semi-supervised K-means \cite{macqueen1965some}, guided by the aligned source domain.
Our significant \textbf{contributions} are, therefore,

 \textbf{[-]} We present a pragmatic problem framework known as AD-GCD, which involves utilizing labeled and unlabeled samples from distinct data distributions in a GCD setup.

 \textbf{[-]} Our solution, \textsc{CDAD-Net}, introduces three key innovations. Our unique domain alignment approach maintains the intrinsic clustering of the target domain. For feature discriminability, we unveil a neighborhood-focused contrastive self-supervision for the target domain. Lastly, we propose a conditional image inpainting task to enhance fine-grained feature association within clusters.

 \textbf{[-]} We establish the experimental setup for AD-GCD on three datasets, and thoroughly analyze \textsc{CDAD-Net} both for cross-domain and in-domain GCD tasks.

\section{Related Works}

\noindent \textbf{Class discovery:} Introduced by Han \etal \cite{Han2019LearningTD}, NCD focuses on categorizing unlabeled samples from new categories by building upon the knowledge gained from a predefined set of categories. Early attempts to tackle the NCD challenge include \cite{Hsu2017LearningTC, Hsu2019MulticlassCW}, both employing two models trained on labeled and unlabeled data separately to facilitate general task transfer learning. There exists a number of follow-up works, including \cite{Han2020AutomaticallyDA, Zhao2021NovelVC,zhong2021neighborhood,Fini_2021_ICCV}, to name a few, which consider mostly parametric classification heads on top of a generic feature extractor for solving NCD.

The GCD task \cite{vaze2022generalized} extends beyond NCD by incorporating unlabeled data from both known and novel classes. \cite{Cao2021OpenWorldSL} tackled this by developing an adaptive margin loss to balance intra-class variances, while \cite{vaze2022generalized} employed pre-trained Vision Transformers (ViTs) \cite{dosovitskiy2020image} and a modified K-Means algorithm for a semi-supervised approach. Further innovations by \cite{Zhang2022PromptCALCA} and \cite{pu2023dynamic} introduced contrastive learning with visual prompts and self-supervised learning strategies to enhance GCD. Meanwhile, \cite{GPC} utilized Gaussian mixture models for clustering, while \cite{chiaroni2023parametric} introduced an information-theoretic solution to GCD. Compared to these non-parametric clustering-based approaches, \cite{simgcd} analyzed the parametric methods for GCD. However, all these endeavors grapple with covariate shifts between datasets.

\noindent \textbf{Cross-domain learning:} Navigating multiple visual domains poses significant challenges, primarily due to appearance drifts across these domains \cite{mdl1, mdl2}. There are two key strategies for handling multi-domain data effectively.
The first, Domain Adaptation (DA) \cite{da1, da2, da3, da4}, aims to align a source domain with a target domain in a transductive learning manner. The second, Domain Generalization (DG) \cite{dg2, dbadg, dg4}, focuses on building a robust and adaptable model applicable to novel target domains during inference. Meanwhile, multi-domain learning \cite{mdl1, mdl2, mdl4} aims to develop a model that leverages labeled data from various domains without letting domain-specific features negatively impact the learning process.
Both DA and DG are applicable in closed-set and open-set scenarios \cite{panareda2017open, saito2018open, daml,Bele_2024_WACV,bose2023beyond}. Open-set DA/DG aims to categorize novel target samples into a singular, undefined class. However, in AD-GCD, the emphasis shifts to discerning and organizing the semantic structures within these unlabeled datasets, given partial supervision from a distinct visual domain.

\noindent \textbf{Self-supervised learning:} 
Self-supervision embodies the concept of deriving meaningful representations directly from data, bypassing the reliance on external semantic labels. These techniques can be broadly classified into contrastive learning-based and more conventional approaches. In the realm of traditional self-supervised models, a pretext task is established, such as solving a jigsaw puzzle, predicting image rotations, or performing image inpainting \cite{ssl1, ssl2, ssl3, ssl4}, among others. These pretext tasks impart visual commonsense to the model.

Numerous studies have delved into the efficacy of contrastive loss in the realm of unsupervised representation learning. This exploration includes seminal works such as InfoNCE \cite{oord2018representation}, SimCLR \cite{SimCLR2020a}, SWaV \cite{Swav2020}, MoCo \cite{MoCo2020}, SEER \cite{SEER2021}, to name a few. Here, the idea is to maximize compatibility between different augmentations of the same image while minimizing the same for distinct images. In place of crafting augmented views to create pairs, \cite{dwibedi2021little} proposed to consider the neighbors. Finally, the notion of multi-domain self-supervised learning is less studied, and existing few methods like \cite{feng2019self, kalibhat2022multidomain} focus on the closed-set scenario. 

Existing GCD models generally use a semi-supervised contrastive learning from labeled and unlabeled data for crafting representations. However, this approach falls short for divergent data distributions, leading us to focus on domain-specific metric objectives. Additionally, we propose to bolster the global image representations by ensuring that they accommodate the local image properties well, through the proposed novel conditional inpainting task.

\section{\textsc{CDAD-Net}: Proposed Methodology}

Consider the dataset $\mathcal{D}$, which encompasses a combination of labeled samples (source domain) denoted as $\mathcal{D_L}$ and unlabeled samples (target domain) referred to as $\mathcal{D_U}$. To elaborate further, $\mathcal{D_L}$ comprises pairs of the form $\{(x_i, y_i)\}_{i=1}^{n_l} \in \mathcal{X \times Y_L}$, while $\mathcal{D_U}$ encompasses pairs $\{(x_j, y_j)\}_{j=1}^{n_u} \in \mathcal{X \times Y_U}$, with the constraint that $\mathcal{Y_L \subset Y_U}$. $\mathcal{Y_L}$ corresponds to the set of labels specifically assigned to the \texttt{known} classes, denoted as $\mathcal{C}_{kwn}$, whereas $\mathcal{Y_U}$ represents the complete label set, encompassing both the \texttt{known} classes and the  \texttt{novel} classes $\mathcal{C}_{new}$, denoted collectively as $\mathcal{C}$. As opposed to the original GCD formulation \cite{vaze2022generalized}, we introduce a new constraint in our proposed AD-GCD setting: $\mathcal{P}(\mathcal{D}_{\mathcal{L}}) \neq \mathcal{P}(\mathcal{D}_{\mathcal{U}})$, which means the domain characteristics of $\mathcal{D}_{\mathcal{L}}$ and $\mathcal{D}_{\mathcal{U}}$ are different.

During the training phase, our model solely has access to the labeled data contained in $\mathcal{D_L}$, while the labels of $\mathcal{D}_{\mathcal{U}}$ are not accessible. Our main goal is to perform accurate clustering of the data samples in $\mathcal{D}_{\mathcal{U}}$ during inference.

\subsection{Model architecture and training details}

\begin{figure*}
  \includegraphics[width = \textwidth ]{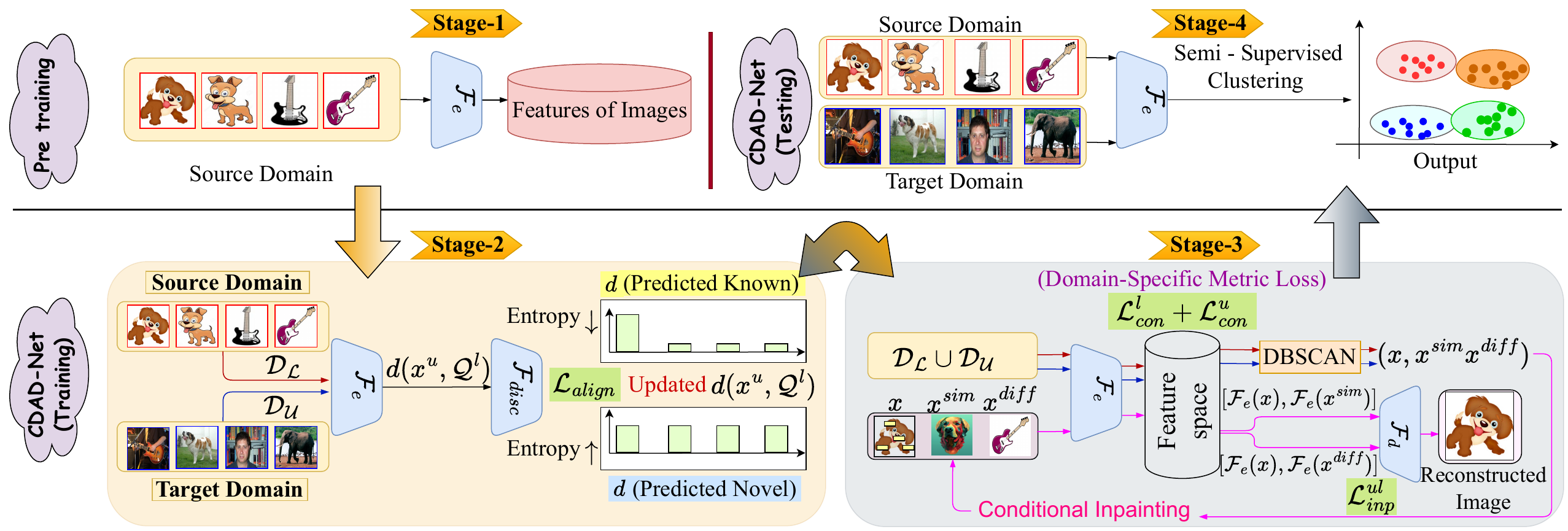} 
  \vspace{-4.5mm}
  \caption{\textbf{Architecture overview and training pipeline of \textsc{CDAD-Net}}. The DINO pre-trained ViT encoder $\mathcal{F}_e$ is first fine-tuned on $\mathcal{D}_{\mathcal{L}}$. Subsequently, the model is trained on $\mathcal{D}_{\mathcal{L}} \cup \mathcal{D}_{\mathcal{U}}$ in two stages in each training epoch, using the domain alignment objective $\mathcal{L}_{align}$ given $(\mathcal{F}_e, \mathcal{F}_{disc})$, and the cumulative metric objective $\mathcal{L}_{con}^l + \mathcal{L}_{con}^u + \mathcal{L}_{inp}^{ul}$ given $(\mathcal{F}_e, \mathcal{F}_d)$. Inference is carried our by semi-supervised K-means applied on $\mathcal{F}_e(\mathcal{D}_{\mathcal{L}}) \cup \mathcal{F}_e(\mathcal{D}_{\mathcal{U}})$. The number of target domain clusters is estimated using the \texttt{elbow} method.}
  \vspace{-5.5mm}
  \label{fig:main}
\end{figure*}

\noindent \textbf{Learning objectives:} As outlined previously, our objective is to address two fundamental challenges of AD-GCD:
\textbf{i)} How to calculate a shared embedding space using the source domain $\mathcal{D}_{\mathcal{L}}$ and the target domain $\mathcal{D}_{\mathcal{U}}$? In this space, we aim to map the known-class samples within the label space $\mathcal{C}_{kwn}$ from $\mathcal{D}_{\mathcal{U}}$ to coincide with $\mathcal{D}_{\mathcal{L}}$, while simultaneously ensuring that novel-class samples from $\mathcal{D}_{\mathcal{U}}$ are positioned distinctively.
\textbf{ii)} How do we ensure the class/cluster discriminativeness for both domains in this space? We seek to optimize the within-class compactness while maximizing the between-class separation given $\mathcal{D}_{\mathcal{L}} \cup \mathcal{D}_{\mathcal{U}}$.
Following the establishment of this shared representation space, we proceed by employing semi-supervised K-Means to cluster the samples contained in $\mathcal{D}_{\mathcal{L}} \cup \mathcal{D}_{\mathcal{U}}$ while ensuring that the labeled data from $\mathcal{D}_{\mathcal{L}}$ and the confident pseudo-labeled data from $\mathcal{D}_{\mathcal{U}}$ preserve their cluster assignments.

\noindent \textbf{Model architectural overview:} Building on the insights from \cite{vaze2022generalized}, we adopt the ViT with a B-16 backbone that has been pre-trained using the DINO objective \cite{dino} on the ImageNet dataset \cite{deng2009imagenet}, as it provides a robust nearest-neighbor classifier, which is deemed to enhance our clustering performance. We denote this pre-trained model as $\mathcal{F}_e$ and use the \texttt{CLS} token for the image embeddings.
As we advance further in our approach, we introduce a discriminator denoted as $\mathcal{F}_{disc}$, which operates on the features extracted by $\mathcal{F}_e$ to oversee the domain alignment objective. In addition to this, we incorporate a decoder network $\mathcal{F}_{d}$ to aid in the conditional image inpainting task at the patch level. 

$\mathcal{F}_d$ denotes the module responsible for inpainting, enacted atop $\mathcal{F}_e$. We infuse the feature encodings from $\mathcal{F}_e$ of the reference image together with that of the companion image without any missing patches to provide the context for reconstruction at different levels of $\mathcal{F}_d$.

\noindent \textbf{Walking through the training of \textsc{CDAD-Net}} (Fig. \ref{fig:main}): 

\textbf{- Fine-tuning $\mathcal{F}_e$ on $\mathcal{D}_{\mathcal{L}}$:} At the beginning, we perform a warm-up supervised fine-tuning of $\mathcal{F}_e$ on $\mathcal{D}_{\mathcal{L}}$.

\textbf{- Proposed alignment of $\mathcal{D}_{\mathcal{L}}$ and $\mathcal{D}_{\mathcal{U}}$:} Given the feature extractor $\mathcal{F}_e$ and the domain discriminator $\mathcal{F}_{disc}$, we propose an entropy-based adversarial learning technique to align the domains, solving our first objective. To elucidate, let's denote $\mathcal{Q}^l = \{\mathcal{Q}_k^l\}_{k=1}^{|\mathcal{C}_{kwn}|}$ as the class prototypes for the source domain classes within $\mathcal{D}_{\mathcal{L}}$. Each $\mathcal{Q}_k^l$ is obtained by averaging the image embeddings given the labeled samples from the $k^{th}$ category. Now, given a sample $x^u$ from $\mathcal{D}_{\mathcal{U}}$, we compute the normalized distance distribution of $x^u$ with respect to $\mathcal{Q}^l$, $d(x^u, \mathcal{Q}^l)$, using the cosine similarity ($\delta$).
\vspace{-1mm}
\begin{equation}
\begin{split}
    d(x^u, \mathcal{Q}^l) = \text{Normalize}[\delta(\mathcal{F}_e(x^u), \mathcal{Q}^l_1), \cdots,\\ \delta(\mathcal{F}_e(x^u), \mathcal{Q}^l_{|\mathcal{C}_{kwn}|})]
\end{split}
\vspace{-1mm}
\end{equation}

 Under this setting, our proposed domain alignment objective focuses on the entropy of $d(x^u, \mathcal{Q}^l)$. In essence, when $x^u$ belongs to one of the known classes in $\mathcal{C}_{kwn}$, we anticipate $d(x^u, \mathcal{Q}^l)$ to exhibit a one-hot type distribution, leading to low entropy. Conversely, suppose $x^u$ originates from one of the novel classes in $\mathcal{C}_{new}$. In that case, we aim for the entropy of $d(x^u, \mathcal{Q}^l)$ to be high, indicating that the sample is distant from the prototypes within $\mathcal{Q}^l$. We represent the corresponding loss objective as $\mathcal{L}_{align}$. 
 
 Existing adversarial OSDA techniques, such as \cite{saito2018open}, utilize a threshold probability of $0.5$ to distinguish unknown samples, typically herding them into a single class label —a practice our proposed method circumvents. Additionally, our feature discrimination tactic, outlined subsequently, enhances cluster density by leveraging $d(x^u, \mathcal{Q}^l)$.

\textbf{- Enforcing discriminativeness objectives:} We introduce two distinct contrastive objectives for both domains alongside a novel image inpainting formulation to improve the cluster density of the learned embeddings. In the source domain, our primary aim is to bring the embeddings of each $x^l$ closer to their respective class prototypes from $\mathcal{Q}^l$ while simultaneously pushing them away from prototypes associated with negative classes. This specific loss is $\mathcal{L}_{con}^l$.
However, achieving this in the target domain presents a significant challenge due to the absence of direct supervision. As a remedy, our method focuses on enhancing contrastive learning by considering near and far neighbors relative to a reference image. To address this, we introduce a technique that leverages the relationships between distances $d(x^{u}_1, \mathcal{Q}^l)$ and $d(x^{u}_2, \mathcal{Q}^l)$. Specifically, if $x^u_1$ and $x^u_2$ are similar, they are deemed to show identical relation with the source domain prototypes.
Accordingly, within a batch, we define a pair $(x_1^u, x_2^u)$ as positive if the Manhattan distance between $d(x^u_1, \mathcal{Q}^l)$ and $d(x^u_2, \mathcal{Q}^l)$ is the smallest among all pairs in the batch. Similarly, we pick $\mathcal{M}$ negative samples based on high Manhattan distance. Subsequently, we apply the contrastive loss on this set, referred to as $\mathcal{L}_{con}^u$. In order to avoid any erroneous positive pairs, we include some image-specific augmentations into our data-pool of the batch.

Moreover, we aim to refine the global \texttt{CLS} embeddings from $\mathcal{F}_e$ to be more responsive to nuanced local image features, a task not directly achieved by $\mathcal{L}_{con}^l + \mathcal{L}_{con}^u$, as they primarily focus to optimize the global image embeddings. To enhance the compactness of our clusters, our strategy is to amplify feature correlations among similar images and reduce them for dissimilar ones, thereby strengthening the clarity and distinction of the clusters.
We approach this through the proposed conditional image inpainting task, using an unmasked full image to assist in the patch-level inpainting of a reference image. 
Precisely, we begin by over-clustering the embeddings obtained from $\mathcal{F}_e(\mathcal{D}_{\mathcal{L}}) \cup \mathcal{F}_e(\mathcal{D}_{\mathcal{U}})$ using DBSCAN \cite{dbscan}, a density-based non-parametric clustering technique.
Within these clusters, we select a quadruplet of samples denoted as $(x, x', x^{sim}, x^{diff})$ where samples in pairs $(x,x^{sim})$ and $(x,x^{diff})$ share similar and different cluster labels, respectively. $x'$ is obtained by removing a random patch from $x$, and we are interested in the inpainting task for $x'$ using $\mathcal{F}_d$. To do this, we enforce $\mathcal{F}_e(x^{sim})$ to provide better assistance than $\mathcal{F}_e(x^{diff})$ since $x$ shares superior semantic similarity to $x^{sim}$ than $x^{diff}$, realised through a novel loss $\mathcal{L}_{inp}^{ul}$.

\subsection{Loss function, training, and inference}

\noindent \textbf{$\mathcal{L}_{align}$: Proposed adversarial domain alignment loss:} To implement the proposed adversarial entropy optimization-based domain alignment method, we introduce the concept of a pseudo-ground-truth for each target domain sample. This pseudo-ground-truth, denoted by $y^g$, is defined as a uniform distribution across $|\mathcal{C}_{kwn}|$ classes. For potential samples from known classes in $\mathcal{D}_{\mathcal{U}}$, our objective is to learn features $\mathcal{F}_e(x^u)$ in such a way that we maximize the cross-entropy loss $\mathcal{L}_{ce}$ between $y^g$ and the distribution $d(x^u, \mathcal{Q}^l)$. 
Conversely, our goal for potential samples from novel classes is to minimize $\mathcal{L}_{ce}$. In this way, we separate the known and novel class samples of $\mathcal{D}_{\mathcal{U}}$ and align the known class samples of both domains.
To achieve our goal, we propose the following loss function:
\vspace{-2mm}
\begin{equation}
    \centering
    \mathcal{L}_{align} = \underset{\mathcal{F}_e}{\min} \ \underset{\mathcal{F}_{disc}}{\max} \underset{\mathcal{P}(\mathcal{D}_{\mathcal{L}} \cup \mathcal{D}_{\mathcal{U}})}{\mathbb{E}} \mathcal{L}_{ce}(y^g, d(x^u, \mathcal{Q}^l)) 
    \vspace{-3mm}
\end{equation}

\noindent \textbf{$\mathcal{L}_{con}^l$: Source domain supervised contrastive loss:} We employ the traditional distance optimization technique when provided with a source domain sample and all the class prototypes in $\mathcal{Q}^l$. Our objective is to maximize the similarity between the sample and prototypes of the same class while minimizing it simultaneously for all negative-class prototypes. This results in the following loss objective, given a $k^{th}$-class sample $x^k$ and class prototype $\mathcal{Q}^l_k$:
\vspace{-2mm}
\begin{equation}
\hspace{-3mm}
\begin{aligned}
\mathcal{L}_{con}^l &= \underset{\mathcal{F}_e}{\min} \ \underset{\mathcal{P}(\mathcal{D}_{\mathcal{L}})}{\mathbb{E}}  -\log \frac{\exp(\delta(\mathcal{F}_e(x^k), \mathcal{Q}_k^l))}{\sum_{\substack{n=1 \\}}^{|\mathcal{C}_{kwn}|}\exp(\delta(\mathcal{F}_e(x^k), \mathcal{Q}_n^l))} 
\end{aligned}
\hspace{-3mm}
\vspace{-1mm}
\end{equation}

\noindent \textbf{$\mathcal{L}_{con}^u$: Target domain contrastive loss with the proposed neighbors selection strategy:} Aligned with our approach in the source domain, we adopt a domain-specific contrastive loss for the target domain, using neighboring sample relationships. The absence of labels for $\mathcal{D}_{\mathcal{U}}$ poses a challenge. To overcome this, we utilize $d(x^u, \mathcal{Q}^l)$ to discern positive from negative target sample pairings for contrastive loss. We strive to enhance the similarity for positive pairs and reduce the same for negative pairs. With an anchor, a positive, and a collective of negative samples, denoted as $(x^u, x^{+u}, \{x^{-u}_m\}_{m=1}^{\mathcal{M}})$, we define $\mathcal{L}_{con}^u$ using the conventional InfoNCE \cite{oord2018representation} based contrastive objective.


\noindent \textbf{$\mathcal{L}_{inp}^{ul}$: Proposed conditional image inpainting loss:} To consolidate our hypothesis that $\mathcal{F}_e(x^{sim})$ should aid in better patch reconstruction of $x'$ than $\mathcal{F}_e(x^{diff})$, we introduce the following novel loss objective, where $\mathcal{L}_{recon}$ defines the inpainting loss. The loss has two components. In order to train $\mathcal{F}_d$ to learn inpainting, we aim to minimize $\mathcal{L}_{recon}(x'|x)$, \ie we condition $\mathcal{F}_d$ on the original reference image itself. 
Besides, we introduce a hinge objective that states if $\mathcal{L}_{recon}(x'|x^{sim}) - \mathcal{L}_{recon}(x'|x^{diff}) \leq 0$, the loss component should return zero, else we seek to minimize the obtained error. As a whole, the combined loss is expected to attain the value zero at optimality, strengthening the fine-grained correlation between $x$ and $x^{sim}$.
\begin{equation}
\begin{split}
    \mathcal{L}_{inp}^{ul} = \min_{\mathcal{F}_e, \mathcal{F}_d} \underset{\mathcal{P}(\mathcal{D}_{\mathcal{L}} \cup \mathcal{D}_{\mathcal{U}})}{\mathbb{E}} \biggl [ \max ( 0,  \mathcal{L}_{recon}(x'|x^{sim}) -  \\ \mathcal{L}_{recon}  (x'|x^{diff}) ) \biggr ] + \mathcal{L}_{recon}(x'|x)
\label{eq:inp}
\end{split}
\end{equation}

\noindent \textbf{Training:} In \textsc{CDAD-Net}, we adopt an iterative optimization technique. First, the warm-up supervised fine-tuning is performed on $\mathcal{F}_e$. Subsequently, each training epoch alternates between $\mathcal{L}_{align}$ for domain alignment and a composite loss $\mathcal{L}_{con}^l + \mathcal{L}_{con}^u + \mathcal{L}_{inp}^{ul}$ for contrastive learning and inpainting. This balanced optimization ensures both domain independence and discriminability in the \texttt{CLS} tokens of $\mathcal{F}_e$.

\noindent \textbf{Inference:}
Consistent with \cite{vaze2022generalized}, we employ $\mathcal{F}_e$'s feature embeddings for applying semi-supervised K-means clustering on $\mathcal{D}_{\mathcal{L}} \cup \mathcal{D}_{\mathcal{U}}$. We ensure accurate categorization for $\mathcal{D}_{\mathcal{L}}$ samples and apply this rigor to confidently pseudo-labeled $\mathcal{D}_{\mathcal{U}}$ samples, preserving cluster consistency and merging supervised insights with clustering. The confidence is calculated based on a sample's \texttt{cosine} similarity to the cluster centre for the known classes. We set the threshold to be 0.9 to avoid any possible misclassification.
For estimating optimal cluster count, we utilize Brent's algorithm \cite{vaze2022generalized}.

\section{Experimental Evaluations}\label{sec:exp}
\label{sec:Experimenatl_protocol}

\begin{table*}[!htbp]
\caption{Comparison of proposed \textsc{CDAD-Net} on AD-GCD with the referred literature (highest in \textcolor{green!20}{green}, second highest in \textcolor{red}{red} and \textbf{bold} is the proposed method)}
\centering

\scalebox{0.8}{
\begin{tabular}{l|ccc|ccc|ccc}
\hline
\multirow{2}{*}{\textbf{Methods}} & \multicolumn{3}{c|}{\textbf{Office-Home}} & \multicolumn{3}{c|}{\textbf{DomainNet}} & \multicolumn{3}{c}{\textbf{PACS}}\\
\cline{2-10}
& \texttt{All} & \texttt{Old} & \texttt{New} & \texttt{All} & \texttt{Old} & \texttt{New} & \texttt{All} & \texttt{Old} & \texttt{New} \\
\hline
\cellcolor{orange!10}K-Means \cite{macqueen1965some} & 29.48  & 32.58 & 24.82 & 24.73 
 & 30.67 & 22.52   & 25.46  & 28.12 & 19.03 \\

\cellcolor{orange!10}ORCA \cite{Cao2021OpenWorldSL} (ICLR'22)&38.23 &40.59 & 29.07  &32.58  & 38.53  & 30.77 &40.62  &43.82 & 35.58    \\


\cellcolor{orange!10}GCD \cite{vaze2022generalized} (CVPR'22)& 50.32 & 54.49 & 45.93 &52.58  & 55.43 &46.12 & 56.71 & 61.66 & 49.32  \\
\cellcolor{orange!10}DCCL \cite{pu2023dynamic} (CVPR'23)& 41.72  & 44.43 & 33.37 & 40.87 & 43.64 & 31.39 & 38.17 & 42.52 & 35.45   \\

\cellcolor{orange!10}GPC \cite{GPC} (ICCV'23)& 42.15  & 43.33 & 30.67 & 39.28 & 44.45 & 29.72 & 36.83 & 44.74 & 31.63  \\

\cellcolor{orange!10}SimGCD \cite{simgcd} (ICCV'23) & 55.67 &60.36 &49.76  &54.27  &65.83  & 41.67 &57.78  &66.82 &48.76\\

\cellcolor{cyan!20}ORCA + OSDA \cite{saito2018open} & 45.25 &47.61 &40.35  &42.38  &45.31  & 39.36 &45.67  &47.21 &44.37 \\

\cellcolor{cyan!20}GPC + OSDA \cite{saito2018open} & 47.84  & 50.62 & 42.16 & 46.52 & 49.46 & 41.84 & 47.53 & 49.14 & 44.84 \\

\cellcolor{cyan!20}GCD + OSDA \cite{saito2018open}  & 61.49 & 65.38 & 53.84 & \cellcolor{red!20}58.79 & \cellcolor{red!20}66.41 &\cellcolor{red!20} 51.28   & 60.52 & 67.69 & 54.50  \\

\cellcolor{cyan!20}DCCL + OSDA \cite{saito2018open} & 52.24  & 52.34 & 45.60 & 52.12 & 54.52   & 43.76 & 58.32 & 60.25 & 55.67 \\

\cellcolor{cyan!20}SimGCD + OSDA \cite{saito2018open} & \cellcolor{red!20}64.04  & \cellcolor{red!20}68.82 & \cellcolor{red!20}58.25 & 51.57 & 52.38  & 43.69 & \cellcolor{red!20}64.83 & \cellcolor{red!20}69.78 & \cellcolor{red!20}60.74  \\
\hline 
\cellcolor{green!20}\textbf{\textsc{CDAD-Net}}  & \cellcolor{green!20}\textbf{67.55} & \cellcolor{green!20}\textbf{72.42} & \cellcolor{green!20}\textbf{63.44}  & \cellcolor{green!20}\textbf{70.28} & \cellcolor{green!20}\textbf{76.46} & \cellcolor{green!20}\textbf{65.19}  & \cellcolor{green!20}\textbf{83.25} & \cellcolor{green!20}\textbf{87.58}& \cellcolor{green!20}\textbf{77.35} \\
\hline
\end{tabular} 
}
\label{tab:cross}
\end{table*}

\noindent \textbf{Datasets:} We run our experiments using six prominent benchmark datasets from the domain adaptation and GCD literature to analyze the cross-domain and within-domain performance of \textsc{CDAD-Net}. Specifically, for AD-GCD, they include:
Office-Home \cite{officehome}, PACS \cite{dbadg}, and DomainNet \cite{domain_net}, respectively. We consider all the pairs of domains to define the source and the target and report the average scores for Office-Home and PACS. For DomainNet, we consider the source-target pairs as listed below:
\vspace{-0.75em}
\begin{enumerate}
    \setlength\itemsep{-0.5em}
    \item Real world $\rightarrow$ Sketch
    \item Painting  $\rightarrow$ Real World
    \item Sketch $\rightarrow$ Clip Art
    \item Sketch $\rightarrow$ Painting
    \item Quickdraw $\rightarrow$ Real World
    \item Sketch $\rightarrow$ Quickdraw
    \item Painting $\rightarrow$ Quickdraw
    \item Painting $\rightarrow$ Infograph
    \item Real World $\rightarrow$ Clip Art
\end{enumerate}
\vspace{-0.75em}
Also, we consider $|\mathcal{C}_{kwn}|:|\mathcal{C}_{new}|$ to be $40:25$, $250:95$, and $4:3$ for Office-Home, DomainNet, and PACS, respectively.
Besides, we show the within-domain class-discovery performance of \textsc{CDAD-Net} on the following datasets:
CIFAR10-100 \cite{cifar10} and
ImageNet100 \cite{deng2009imagenet}, using the established protocols \cite{vaze2022generalized}.\\
\noindent \textbf{Training and implementation details:} In our experimental setup, we initialize $\mathcal{F}_e$ with the pretrained weights of the ViT-B-16 backbone from DINO \cite{dino}. Subsequently, we fine-tune the last block of $\mathcal{F}_e$ with an initial learning rate of 0.01. This learning rate follows a cosine annealed schedule, with $\mathcal{D}_{\mathcal{L}}$ serving as the reference dataset, as outlined in \cite{vaze2022generalized}. To implement the contrastive paradigms, we adhere to the standard practice of using a non-linear projector on top of $\mathcal{F}_e$.
On the other hand, $\mathcal{F}_{disc}$ employs an MLP based architecture. 
We incorporate a random patch masking technique to mask one of the $16$ patches in the original image for the conditional inpainting task. $\mathcal{F}_d$ consists of deconvolution blocks to increase the resolution of the feature maps.  When calculating $\mathcal{L}_{con}^u$, we consider the number of negative samples $\mathcal{M}$ to be $20$.
Regarding DBSCAN clustering, we set the radius value to $1$. This choice intentionally encourages over-clustering, aligning with our goal of creating semantically well-tied local clusters for selecting the triplets.
When estimating the value of $K$ for semi-supervised K-means, we adopt an iterative approach in which we consider $K$ within the range $[|\mathcal{C}_{kwn}|, 1000]$ and employ Brent's algorithm.
Finally, we perform warm-up training for $30$ iterations, followed by main training for $500$ iterations, with a learning rate of $0.01$
using the ADAM \cite{kingma2014adam} optimizer.\\
\noindent \textbf{Evaluation protocols:} 
Our primary evaluation metric, denoted as \texttt{All}, assesses the clustering performance across the entire dataset $\mathcal{D}_{\mathcal{U}}$. In addition to this global evaluation, we also present results for two specific subsets:
\texttt{Old} classes, which consist of instances in $\mathcal{D}_{\mathcal{U}}$ belonging to the $\mathcal{C}_{kwn}$ (known) classes, and
\texttt{New} classes, comprising instances in $\mathcal{D}_{\mathcal{U}}$ that belong to the $\mathcal{C}_{new}$ (new) classes.
We utilize the Hungarian optimal assignment algorithm \cite{kuhn1955hungarian} to compute these subset-specific accuracies to determine the optimal permutation of labels that maximizes accuracy.

\begin{table}[!htbp]
\caption{Comparison of the proposed \textsc{CDAD-Net} to the literature for within-domain GCD task (highest in \textcolor{green!20}{green}, second highest in \textcolor{red}{red} and \textbf{bold} is the proposed method)}
\begin{center}
\scalebox{0.55}{
\begin{tabular}{l|ccc|ccc|ccc}
\hline
\multicolumn{1}{l|}{\multirow{2}{*}{\textbf{Methods}}}&\multicolumn{3}{c|}{\textbf{CIFAR10}} &\multicolumn{3}{c|}{\textbf{CIFAR100}}&\multicolumn{3}{c}{\textbf{ImageNet-100}}\\
\cline{2-10} 
&\texttt{All}&\texttt{Old}&\texttt{New}&\texttt{All}&\texttt{Old}&\texttt{New}&\texttt{All}&\texttt{Old}&\texttt{New}\\
\hline

k-means\cite{macqueen1965some}   &83.60 &85.70 &82.50 &52.00 &52.20 &50.80 &72.70 &75.50 &71.30\\
RankStats+ \cite{rankstat} & 46.80 & 19.20 & 60.50 & 58.20 & 77.60 & 19.30 & 37.10 & 61.60 & 24.80 \\
UNO+ \cite{Fini_2021_ICCV} & 68.60 & \cellcolor{green!20}98.30 & 53.80 & 69.50 & 80.60 & 47.20 & 70.30 & \cellcolor{green!20}95.00 & 57.90\\
ORCA \cite{Cao2021OpenWorldSL} (ICLR'22)&88.90& 88.20& 89.20& 55.10& 65.50& 34.40& 67.60& 9\cellcolor{red!20}0.90& 56.00\\
GCD \cite{vaze2022generalized} (CVPR'22)&91.50 &\cellcolor{red!20}97.90  &88.20 &73.20  &76.20 &66.53 &74.10 &89.80 &66.30 \\
DCCL \cite{pu2023dynamic} (CVPR'23)&\cellcolor{red!20}96.30  & 96.50 & \cellcolor{green!20}96.90 & \cellcolor{red!20}75.30 & \cellcolor{green!20}76.80 &\cellcolor{red!20}70.20 & \cellcolor{red!20}80.50 &90.50 &\cellcolor{red!20}76.20 \\

\hline

\textbf{\textsc{CDAD-Net}} &\cellcolor{green!20}\textbf{96.50} &\textbf{97.20}  &\cellcolor{red!20}\textbf{96.00} &\cellcolor{green!20}\textbf{75.6} &\cellcolor{red!20}\textbf{76.29} &\cellcolor{green!20}\textbf{70.95} &\cellcolor{green!20}\textbf{80.52} &\textbf{80.76} &\cellcolor{green!20}\textbf{80.28} \\

\hline
\end{tabular}}
\label{tab:within}
\end{center}
\end{table}
\begin{figure*}
    \centering
   \includegraphics[scale=0.7]{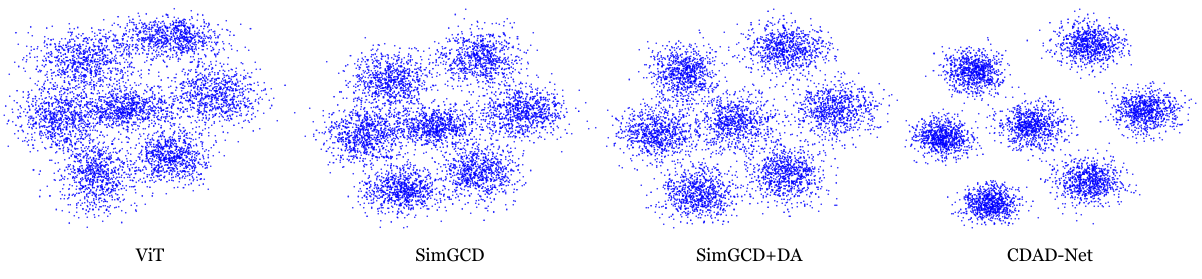}
    \caption{\textbf{t-SNE visualizations} of the target domain clusters, as produced by pre-trained ViT, SimGCD \cite{simgcd}, SimGCD with OSDA \cite{saito2018open}, and \textsc{CDAD-Net} for the PACS dataset, clearly demonstrate that \textsc{CDAD-Net} excels in generating the most distinctive embedding space.}
    
    \label{fig:tsne}
\end{figure*}

\subsection{Comparison to the state-of-the-art} 

\noindent \textbf{Competitors:} We conduct a comparative analysis of \textsc{CDAD-Net} against several existing GCD methods found in the literature. These methods include the following: Baseline K-means \cite{macqueen1965some}, ORCA \cite{Cao2021OpenWorldSL},
GCD \cite{vaze2022generalized}, DCCL \cite{pu2023dynamic}, SimGCD \cite{simgcd}, and GPC \cite{GPC}, respectively.
Furthermore, to address domain shifts effectively, we explore extensions of the GCD, GPC, SimGCD, and DCCL methods where we integrate these techniques with the off-the-shelf OSDA method, OSDA-BP, which is based on adversarial learning \cite{saito2018open}. We adapt the available codebase, and implement DCCL on our own. While integrating OSDA, we follow an alternate optimization strategy between the OSDA and the main objectives. We consider the default parameterization for all these methods.

\noindent \textbf{Discussions on cross-domain GCD performance:} Table \ref{tab:cross} presents a comprehensive comparison of AD-GCD performance across the Office-Home, DomainNet, and PACS datasets when dealing with domains stemming from different data distributions. \textsc{CDAD-Net} consistently outperforms other methods for all domain pairs within these datasets. We performed the experiments on all the domain pairs in OfficeHome and PACS datasets and took the average across all the pairs.

In the case of the Office-Home, \textsc{CDAD-Net} achieves an impressive \texttt{All} accuracy of $67.55\%$, surpassing DCCL by $25.83\%$, SimGCD by $11.88 \%$, GPC by $25.40\%$, and GCD by $17.32\%$.
For the DomainNet , \textsc{CDAD-Net} attains an \texttt{All} accuracy of $70.28\%$, outperforming GCD by $17.7\%$, SimGCD by $16.01 \%$, and GPC by $31\%$.
Finally, in the PACS, \textsc{CDAD-Net} showcases its prowess with an \texttt{All} accuracy of $83.25\%$, surpassing GCD by $26.54\%$, SimGCD by $25.47 \%$, and GPC by $46.42\%$, respectively. 
Moreover, the inclusion of OSDA-BP \cite{saito2018open} did not prove beneficial when integrated with the existing models, given the inherent bias of the OSDA model to cluster the open samples. In this regard, SimGCD+OSDA-BP shows the best performance among the competitors but lags \textsc{CDAD-Net} by $3-18 \%$ in \texttt{All} values.
In contrast, \textsc{CDAD-Net} effectively combines domain alignment with cluster topology preservation for novel classes, leading to improved clustering performance. Qualitatively, we showcase the t-SNE \cite{tsne} visualizations in Fig. \ref{fig:tsne}, confirming that \textsc{CDAD-Net} can produce the most discriminative target domain clustering.

\noindent \textbf{Discussions on within-domain GCD performances:} We present a performance comparison of \textsc{CDAD-Net} with several existing literature on within-domain GCD datasets, as shown in Table \ref{tab:within}.
Notably, our method demonstrates significant superiority over competitors, consistently outperforming them in all the metrics, although the competitors are explicitly designed to handle within-domain GCD.

\begin{table}[!htbp]
\vspace{-2mm}
\caption{Ablation analysis of the proposed model components of \textsc{CDAD-Net} on the OfficeHome dataset.}
\begin{center}
\scalebox{0.6}{
\begin{tabular}{l|c|c|c|c|c|ccc}
\hline
\multirow{2}{*}{\textbf{Conf.}} & \multirow{2}{*}{\thead{Fine-tuned\\ ViT}}  &\multirow{2}{*}{$\mathcal{L}_{con}^l$} & \multirow{2}{*}{$\mathcal{L}_{con}^u$} &\multirow{2}{*}{ $\mathcal{L}_{inp}^{ul}$}
& \multirow{2}{*}{$\mathcal{L}_{align}$}
& \multicolumn{3}{c}{\textbf{Office-Home}}
\\ \cline{7-9} 
 &  &  & &   & &  \texttt{All}  & \texttt{Old} & \texttt{New} \\ \hline
C-1 & $\times$ & $\times$ & $\times$ & $\times$ & $\times$ &66.87 & 69.94 & 60.48 \\
C-2 & $\checkmark$ & $\times$ & $\times$ & $\times$ & $\times$ &67.60 &68.90 &65.35 \\
C-3 & $\checkmark$ &$\checkmark$& $\times$ & $\times$  & $\times$
&72.12 &76.64  &64.87 \\
C-4 & $\checkmark$ &$\checkmark$& $\checkmark$  &$\times$  & $\times$
&72.15 & 76.05 & 65.29\\ 
C-5 & $\checkmark$ &$\checkmark$ &$\checkmark$ & $\checkmark$ & $\times$ &72.93 &76.46 & 66.61 \\
C-6 & $\checkmark$ &$\checkmark$ &$\checkmark$ &$\checkmark$   &$\checkmark$ &\textbf{73.10} &\textbf{75.83}  &\textbf{68.40}\\ \hline
\end{tabular}}
\label{tab:configuration}
\end{center}
\end{table}

\subsection{Ablation analysis}\label{sec:Abalation}
\noindent\textbf{(i) Dissecting the proposed model components:} 
In Table \ref{tab:configuration}, we ablate the performance of various components within \textsc{CDAD-Net} using two domain pairs from the Office-Home dataset, i.e., Art $\rightarrow$ Real World and Clip Art $\rightarrow$ Product.
It is evident that relying solely on source fine-tuned ViT features proves to be suboptimal for achieving effective clustering of $\mathcal{D}_{\mathcal{U}}$. However, the inclusion of the contrastive domain-specific objectives shows further improvements by $\approx 5 \%$ individually in \texttt{All}, while combining them shows some improvements in the \texttt{New} metric. The use of the proposed inpainting objective further improves the performance by $\approx 1.5 \%$ in \texttt{New} metric, signifying its capability to learn more fine-level features into the \texttt{CLS} tokens. The incorporation of $\mathcal{L}^{ul}_{align}$ leads to a notable enhancement, as it assists in domain alignment while not distorting the cluster structure of the classes.. When all the losses are integrated, the model outputs an \texttt{All} value of $73.10 \%$, and \texttt{New} performance of $68.40 \%$.

\noindent \textbf{(ii) Our proposed domain alignment against the existing closed-set and open-set DA strategies:} 
In Table \ref{table:DA1}, we assess the effectiveness of our entropy optimization-based DA strategy in comparison to existing closed-set and open-set DA techniques, specifically DANN \cite{dann} and OSDA-BP \cite{saito2018open} on AD-GCD by integrating these models to \textsc{CDAD-Net}.
While closed-set DA models aim to achieve a global alignment between the two domain distributions, open-set DA techniques focus on isolating outliers to prevent them from affecting the alignment of overlapping classes. However, our observations indicate that both approaches fall short in preserving the cluster structure, particularly on novel classes, thereby diminishing the overall clustering performance.
In contrast, our method does not constrain the variability of the novel class samples, thus less affecting the clustering performance.
\begin{table}[htbp]
\centering
\caption{Comparative analysis of the proposed domain alignment in the \textsc{CDAD-Net} and the state-of-the-art DA approaches.}
\scalebox{0.85}{
\begin{tabular}{l|ccc|ccc}
\hline
 \multicolumn{1}{l|}{\multirow{3}{*}{\textbf{Method}}} &\multicolumn{3}{c|}{\textbf{DomainNet}} &\multicolumn{3}{c}{\textbf{OfficeHome}}\\
\cline{2-7}
&\multicolumn{3}{c|}{Sketch$\rightarrow$Painting} &\multicolumn{3}{c}{Product$\rightarrow$Real World} \\ \cline{2-7}
&\texttt{All} &\texttt{Old}  &\texttt{New} &\texttt{All} &\texttt{Old}  &\texttt{New}  \\ 
\hline
DANN \cite{dann} &57.36 &64.31 &53.44 &36.25 &40.38 &25.67 \\
OSDA \cite{saito2018open} &56.65 & 60.57 & 53.44 &28.85 &31.22 &19.51 \\
\textbf{\textsc{CDAD-Net}} &\textbf{73.42}  &\textbf{76.49} &\textbf{67.70} &\textbf{73.35} &\textbf{79.37} &\textbf{63.08} \\ \hline
\end{tabular}
}
\label{table:DA1}
\end{table}

\begin{figure}[h!]
    \centering
    \includegraphics[scale=0.3]{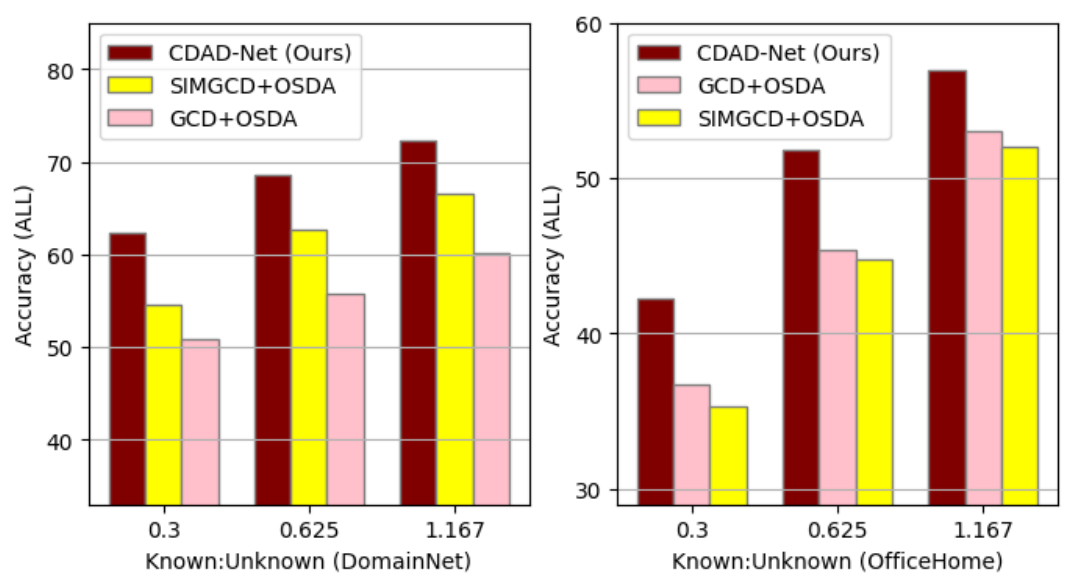}
    
    \caption{Openness analysis of \textsc{CDAD-Net} and competitors on Office-Home and Domain-Net datasets.}
    \label{fig:openness_diff}
\end{figure}
\begin{figure}
    \centering
    \includegraphics[scale=0.2]{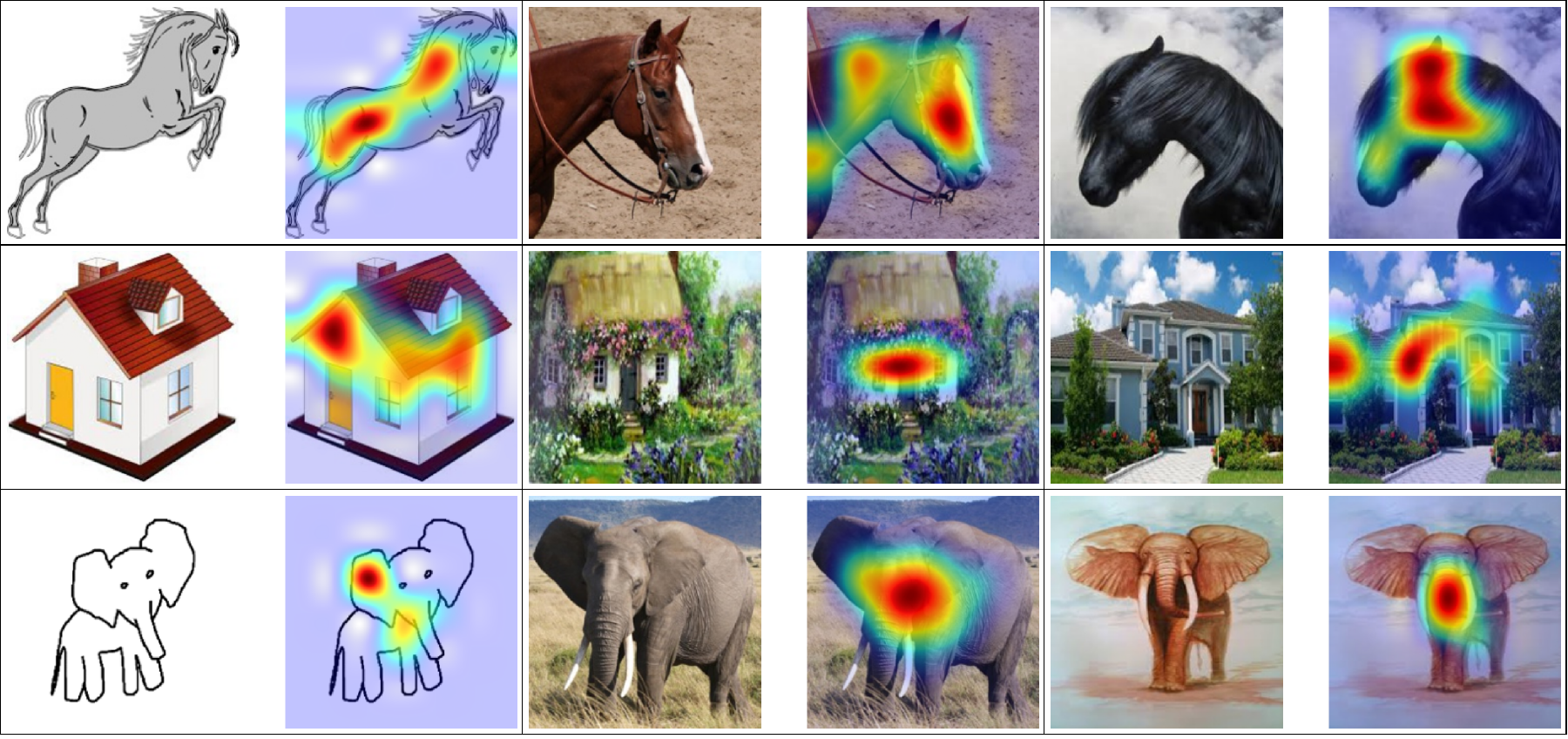}
    \caption{The attention maps produced using GRADCAM.}
    \label{fig:gradcam}
    \end{figure}
\begin{figure}
    \centering
    \includegraphics[scale=0.7]{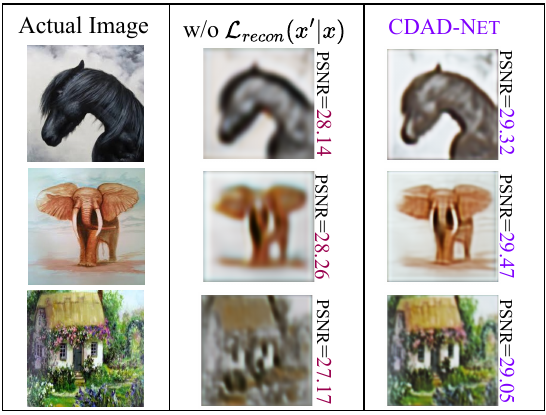}
    
    \vspace{-2mm}
    \caption{Image inpainting outcomes in the absence of $\mathcal{L}_{recon}(x'|x)$ and the full \textsc{CDAD-Net}.}
    \label{fig:inpainting}
\end{figure}

\noindent \textbf{(iii) Sensitivity of \textsc{CDAD-Net} to the number of novel categories in $\mathcal{D}_{\mathcal{U}}$:} We perform a sensitivity analysis of \textsc{CDAD-Net} regarding the number of novel categories in $\mathcal{D}_{\mathcal{U}}$, compared to the known classes overlapped with $\mathcal{D}_{\mathcal{L}}$ (referred to as the openness factor), as depicted in Fig. \ref{fig:openness_diff}. We evaluate the openness of \textsc{CDAD-Net} using domain pairs from Sketch $\rightarrow$ Painting and from Product $\rightarrow$ Real World for the DomainNet and OfficeHome datasets, respectively.
For all openness values, \textsc{CDAD-Net} consistently surpasses its competitors. As the number of known classes decreases in $\mathcal{D}_{\mathcal{U}}$, performance tends to decrease as well. However, the decline in accuracy is less pronounced in \textsc{CDAD-Net}, as evident from the plots. In contrast, comparative techniques exhibit a significant drop in performance as the number of novel classes increases in $\mathcal{D}_{\mathcal{U}}$.\\
\noindent \textbf{(iv) Qualitative visualizations:}
As outlined in \cite{Selvaraju_2019}, we showcase the GRADCAM outputs to visualize attention maps for diverse images across domains in Fig. \ref{fig:gradcam}. The findings indicate that the model adeptly focuses on the discriminative features crucial for classification. Similarly, we showcase the results of the inpainting with respect to our proposed loss measures in Fig. \ref{fig:inpainting}. It is evident from the achieved PSNR scores that the use of the hinge objective aids in better patch reconstruction. 

\section{Takeaways and Future Scope}

In this paper, we propose the challenging new task of AD-GCD, wherein disparate distributions between labeled and unlabeled samples in the GCD setting can thwart the transfer of insights from labeled to unlabeled sets for successful clustering. We tackle this with our \textsc{CDAD-Net} framework, introducing a targeted domain alignment goal that aligns domains while preserving the integrity of cluster structures across all known and emerging categories. Concurrently, we devise a unique domain-centric metric learning approach, employing contrastive methods to bolster discriminability across domains. We further incorporate a conditional image inpainting mechanism to enrich fine-grained feature associations within image embeddings of similar locales. Our empirical evaluations verify \textsc{CDAD-Net}'s superior performance in cross-domain and within-domain GCD tasks, laying the groundwork for future applications like multi-camera person re-identification and other cross-domain learning-driven applications.

{\small
\bibliographystyle{ieee_fullname}
\bibliography{egbib}
}

\clearpage
\appendix

\section{Contents of the supplementary materials}
In the supplementary material, we provide detailed description of dataset splits, further experimental results, including:
\begin{enumerate}
    \item In Table \ref{tab:data}, we provide the dataset splits for the GCD settings for PACS \cite{dbadg}, OfficeHome \cite{officehome}, DomainNet \cite{domain_net}, CIFAR-10 \cite{cifar10}, CIFAR-100 and ImageNet-100 \cite{deng2009imagenet} datasets.

    \begin{table}[ht!]
    \centering
    \caption{The dataset splits (labeled and unlabeled) for the AD-GCD and within-domain GCD experiments.}
    \scalebox{1}{
    \begin{tabular}{l|ll|ll}
    \hline
        Dataset & $|\mathcal{Y}_{\mathcal{L}}|$ & $|\mathcal{D}_{\mathcal{L}}|$ & $|\mathcal{Y}_{\mathcal{U}}|$ & $|\mathcal{D}_{\mathcal{U}}|$ \\ 
        \hline
        \cellcolor[gray]{0.9}PACS \cite{dbadg}&\cellcolor[gray]{0.9}4& \cellcolor[gray]{0.9}1.4K&\cellcolor[gray]{0.9}7&\cellcolor[gray]{0.9}2.4K\\ 
        \cellcolor[gray]{0.9}Office-Home \cite{officehome} & \cellcolor[gray]{0.9}40 &\cellcolor[gray]{0.9}7K &\cellcolor[gray]{0.9}65 &\cellcolor[gray]{0.9}7.5K \\ 
        \cellcolor[gray]{0.9}DomainNet \cite{domain_net} &\cellcolor[gray]{0.9}250 &\cellcolor[gray]{0.9}365.1K &\cellcolor[gray]{0.9}345 &\cellcolor[gray]{0.9}134.5K \\ 
        \cellcolor{red!20}CIFAR-10 \cite{cifar10} &\cellcolor{red!20}5 & \cellcolor{red!20}12.5K &\cellcolor{red!20}10 &\cellcolor{red!20}37.5K \\ 
        \cellcolor{red!20}CIFAR-100 \cite{cifar10}&\cellcolor{red!20}80 & \cellcolor{red!20}20K &\cellcolor{red!20}100 &\cellcolor{red!20}30K \\ 
        \cellcolor{red!20}ImageNet-100 \cite{deng2009imagenet} &\cellcolor{red!20}50 &\cellcolor{red!20}31.9K &\cellcolor{red!20}100 &\cellcolor{red!20}95.3K \\ 
        \hline
    \end{tabular}}
\label{tab:data}
    
\end{table}
    \item In Figure \ref{fig:diffpatches}, we ablate \textsc{CDAD-Net} by varying the number of patches in the masked input image on the OfficeHome Dataset (Art $\rightarrow$ Real World). We observe that as the number of masked input patches increases the classification performance decreases significantly for \texttt{All}, \texttt{Old} and \texttt{New} classes.
    \begin{figure}[ht!]
  \includegraphics[width = \linewidth ]{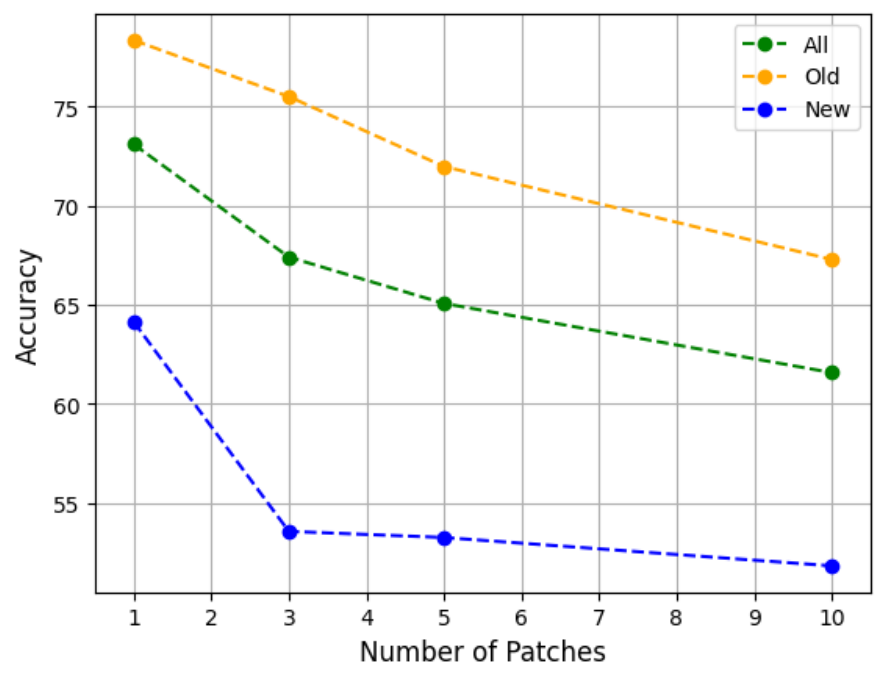}
  \vspace{-4.5mm}
  \caption{Variation of accuracies with the variation in number of patches introduced in the masked input image for the OfficeHome Dataset (Art $\rightarrow$ Real World).}
  \vspace{-5.5mm}
  \label{fig:diffpatches}
\end{figure}
    
    \item In Tables \ref{tab:pacs}, \ref{tab:oh} and \ref{tab:dn}, we perform experiments for all the combinations on the PACS, OfficeHome and DomainNet datasets, respectively. Also, we showcase the list of combinations used in DomainNet dataset in following manner:
   
    \vspace{-0.75em}
    \begin{enumerate}
        \setlength\itemsep{-0.125em}
        \item Real world $\rightarrow$ Sketch
        \item Painting  $\rightarrow$ Real World
        \item Sketch $\rightarrow$ Clip Art
        \item Sketch $\rightarrow$ Painting
        \item Quickdraw $\rightarrow$ Real World
        \item Sketch $\rightarrow$ Quickdraw
        \item Painting $\rightarrow$ Quickdraw
        \item Painting $\rightarrow$ Infograph
        \item Real World $\rightarrow$ Clip Art
    \end{enumerate}
    \vspace{-0.75em}
    In tables, we emphasize our findings using \textbf{bold} text. Additionally, we highlight cells containing the highest and second-highest values with shades of \textcolor{green!20}{green} and \textcolor{red!20}{red}, respectively.

\end{enumerate}

\begin{table*}[ht!]
        \centering
        \caption{Detailed comparison of our proposed \textsc{CDAD-Net} on AD-GCD with respect to the referred literature for the PACS dataset}
        \scalebox{0.71}{
        \begin{tabular}{l|ccc|ccc|ccc|ccc|ccc|ccc}\hline
          \multicolumn{1}{l|}{\multirow{3}{*}{\textbf{Method}}}&	\multicolumn{18}{c}{\textbf{PACS}}\\\cline{2-19}
           &	\multicolumn{3}{c}{Photo $\rightarrow$ Cartoon} &\multicolumn{3}{c}{Art Painting $\rightarrow$ Cartoon}
           &\multicolumn{3}{c}{Art Painting $\rightarrow$ Photo}
           &\multicolumn{3}{c}{Art Painting $\rightarrow$ Sketch}
           &\multicolumn{3}{c}{Cartoon $\rightarrow$ Art Painting}
           &\multicolumn{3}{c}{Cartton $\rightarrow$ Photo}\\\cline{2-19}
            & \texttt{All}& \texttt{Old} & \texttt{New} &  \texttt{All}& \texttt{Old} & \texttt{New}& \texttt{All}& \texttt{Old} & \texttt{New}& \texttt{All}& \texttt{Old} & \texttt{New}& \texttt{All}& \texttt{Old} & \texttt{New}&  \texttt{All}& \texttt{Old} & \texttt{New}\\\hline
\cellcolor{orange!10}GCD \cite{vaze2022generalized}&\cellcolor{red!20}56.44 & 55.26 & \cellcolor{red!20}57.97 & 44.95 & 51.56 & 36.33 &\cellcolor{red!20} 82.97 & \cellcolor{red!20}98.90 &\cellcolor{red!20} 69.66 & 38.50 & 42.40 & 27.88 & 50.65 & 61.60 & 37.86 & \cellcolor{red!20}79.66 & \cellcolor{red!20}93.06 & \cellcolor{red!20}68.49\\
\cellcolor{orange!10}SimGCD	\cite{simgcd}&34.21 & 43.03 & 22.69 & 40.36 & 44.96 & 34.39 & 58.06 & 72.6 & 45.92 & 40.34 &\cellcolor{red!20} 46.9 & 22.36 & 44.35 & \cellcolor{red!20}75.83 & 7.60 & 47.02 & 83.55 & 16.59\\
\cellcolor{cyan!20}GCD+OSDA \cite{saito2018open}&52.90 & \cellcolor{red!20}58.25 & 45.93 & \cellcolor{red!20}52.63 & \cellcolor{red!20}54.09 & \cellcolor{red!20}50.72 & 75.33 & 97.2 & 57.04 & \cellcolor{red!20}42.86 & 42.24 & 44.53 & \cellcolor{red!20}52.54 & 64.64 & \cellcolor{red!20}38.43 & 66.31 & 87.83 & 48.38\\
\cellcolor{cyan!20}SimGCD+OSDA	\cite{saito2018open}&34.13 & 44.85 & 20.16 & 20.03 & 34.09 & 11.68 & 55.99 & 73.10 & 41.70 & 35.10 & 31.34 & \cellcolor{red!20}45.32 & 44.31 & 70.96 & 13.20 & 44.97 & 60.15 & 32.30\\
\cellcolor{blue!20}\textbf{\textsc{CDAD-Net}}	&\cellcolor{green!20}\textbf{69.03} & \cellcolor{green!20}\textbf{63.45} & \cellcolor{green!20}\textbf{76.30} & \cellcolor{green!20}\textbf{70.82} & \cellcolor{green!20}\textbf{68.65} & \cellcolor{green!20}\textbf{73.65} & \cellcolor{green!20}\textbf{99.40} & \cellcolor{green!20}\textbf{99.60} & \cellcolor{green!20}\textbf{99.23} & \cellcolor{green!20}\textbf{52.05} & \cellcolor{green!20}\textbf{52.38} & \cellcolor{green!20}\textbf{51.14} & \cellcolor{green!20}\textbf{92.09} & \cellcolor{green!20}\textbf{89.03} & \cellcolor{green!20}\textbf{95.66} & \cellcolor{green!20}\textbf{99.28} & \cellcolor{green!20}\textbf{99.34} & \cellcolor{green!20}\textbf{99.23}
\\ \bottomrule

\multicolumn{1}{l|}{\multirow{3}{*}{\textbf{Method}}}&	\multicolumn{18}{c}{\textbf{PACS}}\\\cline{2-19}
           &	\multicolumn{3}{c}{Cartoon $\rightarrow$ Sketch} &\multicolumn{3}{c}{Photo $\rightarrow$ Art Painting}
           &\multicolumn{3}{c}{Photo $\rightarrow$ Sketch}
           &\multicolumn{3}{c}{Sketch $\rightarrow$ Art Painting}
           &\multicolumn{3}{c}{Sketch $\rightarrow$ Cartoon}
           &\multicolumn{3}{c}{Sketch $\rightarrow$ Photo}\\\cline{2-19}
            & \texttt{All}& \texttt{Old} & \texttt{New} &  \texttt{All}& \texttt{Old} & \texttt{New}& \texttt{All}& \texttt{Old} & \texttt{New}& \texttt{All}& \texttt{Old} & \texttt{New}& \texttt{All}& \texttt{Old} & \texttt{New}&  \texttt{All}& \texttt{Old} & \texttt{New}\\\hline
\cellcolor{orange!10}GCD \cite{vaze2022generalized}&41.38 & 42.42 & 38.58 &\cellcolor{red!20} 80.41 & 79.61 & \cellcolor{red!20}81.34 & \cellcolor{red!20}43.80 & \cellcolor{red!20}40.11 & \cellcolor{red!20}54.10 & 34.07 & 40.06 & 27.08 &\cellcolor{red!20} 43.55 & 51.69 & 32.93 &\cellcolor{red!20} 54.90 & \cellcolor{red!20}58.01 & 52.32\\
\cellcolor{orange!10}SimGCD	\cite{simgcd}&42.75 & \cellcolor{red!20}51.42 & 19.15 & 49.62 & \cellcolor{red!20}87.26 & 5.69 & 30.86 & 37.68 & 12.31 & 29.90 & \cellcolor{red!20}48.80 & 7.97 & 40.14 & \cellcolor{red!20}68.44 & 3.23 & 36.20 & 29.24 & 42.00\\
\cellcolor{cyan!20}GCD+OSDA \cite{saito2018open}&\cellcolor{red!20}43.93 & 45.14 & \cellcolor{red!20}40.64 & 60.90 & 75.68 & 43.81 & 34.70 & 33.42 & 38.18 & \cellcolor{red!20}34.27 & 32.01 & 36.90 & 38.76 & 40.19 & \cellcolor{red!20}36.90 & 45.51 & 35.92 & \cellcolor{red!20}53.50\\
\cellcolor{cyan!20}SimGCD+OSDA	\cite{saito2018open}&35.49 & 40.31 & 22.37 & 51.51 & 80.84 & 17.27 & 23.23 & 17.30 & 39.36 & 24.00 & 11.51 &\cellcolor{red!20} 38.58 & 19.96 & 34.43 & 10.20 & 26.71 & 11.50 & 47.65\\
\cellcolor{blue!20}\textbf{\textsc{CDAD-Net}}&\cellcolor{green!20}\textbf{51.74} & \cellcolor{green!20}\textbf{53.01} & \cellcolor{green!20}\textbf{48.30} & \cellcolor{green!20}\textbf{92.30} & \cellcolor{green!20}\textbf{90.10} & \cellcolor{green!20}\textbf{95.60} & \cellcolor{green!20}\textbf{49.66} & \cellcolor{green!20}\textbf{43.30} & \cellcolor{green!20}\textbf{56.95} & \cellcolor{green!20}\textbf{90.53} & \cellcolor{green!20}\textbf{86.67} & \cellcolor{green!20}\textbf{95.03} & \cellcolor{green!20}\textbf{72.10} & \cellcolor{green!20}\textbf{73.55} & \cellcolor{green!20}\textbf{70.21} & \cellcolor{green!20}\textbf{99.34} & \cellcolor{green!20}\textbf{99.47} & \cellcolor{green!20}\textbf{99.23}
\\ \bottomrule
            
        \end{tabular} 
        }
    \label{tab:pacs}
\end{table*}

\begin{table*}[ht!]
        \centering
        \caption{Detailed comparison of our proposed \textsc{CDAD-Net} on AD-GCD with respect to the referred literature for the OfficeHome dataset}
        \scalebox{0.71}{
        \begin{tabular}{l|ccc|ccc|ccc|ccc|ccc|ccc}\hline
          \multicolumn{1}{l|}{\multirow{3}{*}{\textbf{Method}}}&	\multicolumn{18}{c}{\textbf{OfficeHome}}\\\cline{2-19}
           &	\multicolumn{3}{c}{Real World $\rightarrow$ Clipart} &\multicolumn{3}{c}{Product $\rightarrow$ Real World}
           &\multicolumn{3}{c}{Art $\rightarrow$ Clipart}
           &\multicolumn{3}{c}{Art $\rightarrow$ Product}
           &\multicolumn{3}{c}{Art $\rightarrow$ Real World}
           &\multicolumn{3}{c}{Clipart $\rightarrow$ Art}\\\cline{2-19}
            & \texttt{All}& \texttt{Old} & \texttt{New} &  \texttt{All}& \texttt{Old} & \texttt{New}& \texttt{All}& \texttt{Old} & \texttt{New}& \texttt{All}& \texttt{Old} & \texttt{New}& \texttt{All}& \texttt{Old} & \texttt{New}&  \texttt{All}& \texttt{Old} & \texttt{New}\\\hline
\cellcolor{orange!10}GCD \cite{vaze2022generalized}&	26.49 & 39.47 & 28.38 & 62.42 & 77.32 & 60.87 & 27.01 & 39.94 & 28.96 & 64.04 & 72.32 & 62.04 & 64.93 & 75.98 & 58.86 & 52.25 & 62.19 & 44.75\\
\cellcolor{orange!10}SimGCD	\cite{simgcd}&40.22 & 48.33 &\cellcolor{red!20}46.12 & 67.47 & 84.32 & 37.80 & 41.57 & 49.23 & \cellcolor{red!20}47.19 & 59.78 & 75.00 & 58.69 & 75.89 & 86.55 & 60.37 & 57.10 & 66.79 & \cellcolor{red!20}49.93\\
\cellcolor{cyan!20}GCD+OSDA \cite{saito2018open}&44.54 & 52.08 & 40.18 & \cellcolor{red!20}70.51 & 89.44 & 59.02 & 46.21 & 53.24 & 44.38 &\cellcolor{red!20} 73.76 & \cellcolor{red!20}81.88 & \cellcolor{red!20}68.73 & 75.55 & 87.42 & 65.29 & 61.88 & 72.26 & 48.90\\
\cellcolor{cyan!20}SimGCD+OSDA	\cite{saito2018open}&\cellcolor{green!20}53.16 & \cellcolor{red!20}56.63 & \cellcolor{green!20}48.73 & 38.31 & \cellcolor{red!20}93.29 & \cellcolor{green!20}77.70 & \cellcolor{green!20}53.96 &\cellcolor{red!20} 57.23 & \cellcolor{green!20}49.86 & 72.71 & 81.57 & 55.72 & \cellcolor{red!20}77.62 & \cellcolor{red!20}93.77 & \cellcolor{red!20}71.09 & \cellcolor{green!20}71.09 & \cellcolor{green!20}78.58 & 49.39\\
\cellcolor{blue!20}\textbf{\textsc{CDAD-Net}}	&\cellcolor{red!20}\textbf{49.52} & \cellcolor{green!20}\textbf{62.44} & \textbf{42.91} & \cellcolor{green!20}\textbf{83.28} & \cellcolor{green!20}\textbf{95.64} & \cellcolor{red!20}\textbf{75.64} & \cellcolor{red!20}\cellcolor{red!20}\textbf{48.46} & \cellcolor{green!20}\textbf{60.10} & \textbf{44.01} & \cellcolor{green!20}\textbf{83.05} & \cellcolor{green!20}\textbf{92.58} & \cellcolor{green!20}\textbf{81.74} & \cellcolor{green!20}\textbf{83.03} & \cellcolor{green!20}\textbf{95.69} & \cellcolor{green!20}\textbf{76.70} & \cellcolor{red!20}\textbf{66.25} & \cellcolor{red!20}\textbf{76.59} & \cellcolor{green!20}\textbf{62.22}
\\ \bottomrule

\multicolumn{1}{l|}{\multirow{3}{*}{\textbf{Method}}}&	\multicolumn{18}{c}{\textbf{OfficeHome}}\\\cline{2-19}
           &	\multicolumn{3}{c}{Clipart $\rightarrow$ Product} &\multicolumn{3}{c}{Clipart $\rightarrow$ Real World}
           &\multicolumn{3}{c}{Product $\rightarrow$ Art}
           &\multicolumn{3}{c}{Product $\rightarrow$ Clipart}
           &\multicolumn{3}{c}{Real World $\rightarrow$ Art}
           &\multicolumn{3}{c}{Real World $\rightarrow$ Clipart}\\\cline{2-19}
            & \texttt{All}& \texttt{Old} & \texttt{New} &  \texttt{All}& \texttt{Old} & \texttt{New}& \texttt{All}& \texttt{Old} & \texttt{New}& \texttt{All}& \texttt{Old} & \texttt{New}& \texttt{All}& \texttt{Old} & \texttt{New}&  \texttt{All}& \texttt{Old} & \texttt{New}\\\hline
\cellcolor{orange!10}GCD \cite{vaze2022generalized}&63.26 & 72.72 & 59.08 & 62.07 & 74.09 & 67.86 & 53.92 & 62.72 & 49.00 & 29.28 & 37.73 & 27.78 & 55.40 & 66.08 & 46.22 & 30.55 & 39.47 & 28.27\\
\cellcolor{orange!10}SimGCD	\cite{simgcd}&57.03 & 72.65 & 47.50 & 60.81 & 78.95 & 70.42 & 60.84 & 69.57 & 45.64 & 39.22 & 43.12 & \cellcolor{red!20}50.98 & 67.62 & 81.31 & 50.88 & 40.49 & 46.51 & 38.81\\
\cellcolor{cyan!20}GCD+OSDA \cite{saito2018open}&\cellcolor{red!20}70.69 & \cellcolor{red!20}79.25 & \cellcolor{red!20}64.93 & 69.18 & 75.68 &\cellcolor{red!20} 77.29 & 65.39 & 75.17 & 53.79 & 43.48 & 51.26 & 40.42 & 77.45 & 83.93 & 53.71 & 45.14 & 53.43 & 41.21\\
\cellcolor{cyan!20}SimGCD+OSDA	\cite{saito2018open}&69.35 & 79.03 & 60.84 &\cellcolor{red!20} 71.99 & \cellcolor{red!20}79.32 & 70.30 &\cellcolor{red!20} 74.20 & \cellcolor{green!20}81.57 & \cellcolor{red!20}62.77 & \cellcolor{green!20}52.85 & \cellcolor{red!20}54.75 & \cellcolor{green!20}61.05 & \cellcolor{green!20}86.03 & \cellcolor{green!20}89.47 & \cellcolor{green!20}63.88 & \cellcolor{green!20}53.1 &\cellcolor{red!20} 56.31 & \cellcolor{green!20}49.11\\
\cellcolor{blue!20}\textbf{\textsc{CDAD-Net}}	&\cellcolor{green!20}\textbf{82.96} & \cellcolor{green!20}\textbf{91.47} & \cellcolor{green!20}\textbf{85.09} & \cellcolor{green!20}\textbf{80.51} & \cellcolor{green!20}\textbf{84.86} & \cellcolor{green!20}\textbf{83.14} & \cellcolor{green!20}\textbf{69.10} & \cellcolor{red!20}\textbf{79.85} & \cellcolor{green!20}\textbf{64.11} &\cellcolor{red!20} \textbf{43.49} & \cellcolor{green!20}\textbf{55.54} & \textbf{39.05} & \cellcolor{red!20}\textbf{78.23} &\cellcolor{red!20} \textbf{85.37} & \cellcolor{red!20}\textbf{61.82} & \cellcolor{red!20}\textbf{48.62} & \cellcolor{green!20}\textbf{61.27} & \cellcolor{red!20}\textbf{42.48}
\\ \bottomrule
            
        \end{tabular}
        }
    \label{tab:oh}
\end{table*}
  
\begin{table*}[ht!]
        \centering
        \caption{Detailed comparison of our proposed \textsc{CDAD-Net} on AD-GCD with respect to the referred literature for the DomainNet dataset }
        \scalebox{0.71}{
        \begin{tabular}{l|ccc|ccc|ccc|ccc|ccc}\hline
          \multicolumn{1}{l|}{\multirow{3}{*}{\textbf{Method}}}&	\multicolumn{15}{c}{\textbf{DomainNet}}\\\cline{2-16}
           &	\multicolumn{3}{c}{Real World $\rightarrow$ Sketch} &\multicolumn{3}{c}{Painting $\rightarrow$ Real World}
           &\multicolumn{3}{c}{Sketch $\rightarrow$ Clipart}
           &\multicolumn{3}{c}{Sketch $\rightarrow$ Painting}
           &\multicolumn{3}{c}{Quickdraw $\rightarrow$ Real World}\\\cline{2-16}
            & \texttt{All}& \texttt{Old} & \texttt{New} &  \texttt{All}& \texttt{Old} & \texttt{New}& \texttt{All}& \texttt{Old} & \texttt{New}& \texttt{All}& \texttt{Old} & \texttt{New}& \texttt{All}& \texttt{Old} & \texttt{New}\\\hline
\cellcolor{orange!10}GCD \cite{vaze2022generalized}&26.17 & 34.09 & 22.65 & 35.21 & 39.94 & 27.27 & 25.21 & 39.94 & 11.27 & 44.44 & 48.67 & 30.52 & 29.30 & 35.98 & 22.17\\
\cellcolor{orange!10}SimGCD	\cite{simgcd}&29.93 & 42.70 & 25.60 & 36.97 &\cellcolor{green!20} 48.98 & 32.47 & 29.69 & 44.98 & 17.89 & 46.71 & 48.57 & 36.12 & 31.62 & 39.44 & 26.88	\\
\cellcolor{cyan!20}GCD+OSDA \cite{saito2018open}&45.71 & \cellcolor{red!20}50.60 & 38.39 & 39.36 & 41.66 & 34.70 & 32.74 & 47.66 & 19.80 & 583.15 & 58.29 & 52.15 & 35.38 & 45.84 & 29.71\\
\cellcolor{cyan!20}SimGCD+OSDA	\cite{saito2018open}&4\cellcolor{red!20}8.16 & 50.32 & \cellcolor{red!20}45.46 & \cellcolor{red!20}41.34 & 42.01 &\cellcolor{red!20} 36.80 & \cellcolor{red!20}35.97 & \cellcolor{red!20}\cellcolor{red!20}49.34 & \cellcolor{red!20}21.56 & \cellcolor{red!20}60.72 & \cellcolor{red!20}65.68 & \cellcolor{red!20}58.42 & \cellcolor{red!20}39.63 & \cellcolor{red!20}47.88 & \cellcolor{red!20}32.79\\
\cellcolor{blue!20}\textbf{\textsc{CDAD-Net}}	&\cellcolor{green!20}\textbf{58.07} & \cellcolor{green!20}\textbf{63.51} & \cellcolor{green!20}\textbf{55.79} & \cellcolor{green!20}\textbf{45.74} & \cellcolor{red!20}\textbf{42.15} & \cellcolor{green!20}\textbf{38.56} & \cellcolor{green!20}\textbf{39.68} & \cellcolor{green!20}\textbf{51.15} & \cellcolor{green!20}\textbf{27.64} & \cellcolor{green!20}\textbf{73.42} & \cellcolor{green!20}\textbf{76.49} & \cellcolor{green!20}\textbf{67.70} & \cellcolor{green!20}\textbf{43.40} & \cellcolor{green!20}\textbf{51.28} & \cellcolor{green!20}\textbf{38.34}
\\ 
\cline{1-16}
\multicolumn{1}{l|}{\multirow{3}{*}{\textbf{Method}}}&	\multicolumn{12}{c}{\textbf{DomainNet}}\\\cline{2-13}
           &	\multicolumn{3}{c}{Sketch $\rightarrow$ Quickdraw} &\multicolumn{3}{c}{Painting $\rightarrow$ Quickdraw}
           &\multicolumn{3}{c}{Painting $\rightarrow$ Infograph}
           &\multicolumn{3}{c}{Real World $\rightarrow$ Clipart}\\\cline{2-13}
            & \texttt{All}& \texttt{Old} & \texttt{New} &  \texttt{All}& \texttt{Old} & \texttt{New}& \texttt{All}& \texttt{Old} & \texttt{New}& \texttt{All}& \texttt{Old} & \texttt{New}\\\cline{2-13}
\cellcolor{orange!10}GCD \cite{vaze2022generalized}&26.35 & 32.19 & 19.64 & 27.36 & 32.72 & 17.39 & 16.17 & \cellcolor{red!20}24.09 & 12.65 & 35.62 & 39.49 & 31.79	\\
\cellcolor{orange!10}SimGCD	\cite{simgcd}&24.22 & 38.39 & 21.44 & 24.15 & 35.41 & 19.56 & \cellcolor{red!20}20.93 & 22.70 & \cellcolor{red!20}15.60 & 39.21 & 45.94 & 41.27\\
\cellcolor{cyan!20}GCD+OSDA \cite{saito2018open}&29.41 & 40.83 & 25.32 & \cellcolor{red!20}47.67 & \cellcolor{green!20}60.22 & \cellcolor{red!20}44.35 & 18.71 & 20.60 & 13.39 & 45.74 & \cellcolor{red!20}53.67 & 35.68\\
\cellcolor{cyan!20}SimGCD+OSDA	\cite{saito2018open}&\cellcolor{red!20}35.10 & \cellcolor{red!20}46.97 & \cellcolor{red!20}37.09 & 42.14 & 43.36 & 36.54 & 16.00 & 20.32 & 9.46 & \cellcolor{red!20}47.97 & 53.34 & \cellcolor{red!20}45.56\\
\cellcolor{blue!20}\textbf{\textsc{CDAD-Net}}	&
\cellcolor{green!20}\textbf{57.23} & \cellcolor{green!20}\textbf{52.58} & \cellcolor{green!20}\textbf{46.83} &\cellcolor{green!20}\textbf{52.49} & \cellcolor{red!20}\textbf{54.48} & \cellcolor{green!20}\textbf{48.79} & \cellcolor{green!20}\textbf{26.07} & \cellcolor{green!20}\textbf{29.51} & \cellcolor{green!20}\textbf{15.79} & \cellcolor{green!20}\textbf{60.28} & \cellcolor{green!20}\textbf{67.15} & \cellcolor{green!20}\textbf{58.42}
\\ \cline{2-13}
            
        \end{tabular} 

        }
    \label{tab:dn}
\end{table*}
\end{document}